# Modal Logics for Qualitative Possibility and Beliefs

**Craig Boutilier**
Department of Computer Science
University of British Columbia
Vancouver, British Columbia
CANADA, V6T 1Z2
email: cebly@cs.ubc.ca

## Abstract

Possibilistic logic has been proposed as a numerical formalism for reasoning with uncertainty. There has been interest in developing qualitative accounts of possibility, as well as an explanation of the relationship between possibility and modal logics. We present two modal logics that can be used to represent and reason with qualitative statements of possibility and necessity. Within this modal framework, we are able to identify interesting relationships between possibilistic logic, beliefs and conditionals. In particular, the most natural conditional definable via possibilistic means for default reasoning is identical to Pearl's conditional for $\varepsilon$-semantics.

## 1 Introduction

There has been a great deal of interest in the relationship between numeric and non-numeric approaches to uncertain reasoning. Possibilistic has been proposed as one such numeric formalism (see (Dubois and Prade 1988) for and introduction), providing *necessity measures*, which determine the degree of certainty associated with an item of belief, and the dual *possibility measures*, determining the degree of surprise associated with (or the willingness to accept) a potential belief. Naturally, qualitative accounts of possibilistic logic have been proposed and shown to correspond to these measures (Dubois 1986). Such qualitative characterizations give us the ability to express possibilistic relationships without having to assume particular numerical values, relying only on the *relative* possibility of propositions.

A logic of qualitative possibility is crucial if we wish to derive consequences based on partial information. Given some constraints on the relationship between certain propositions, certain other constraints may be required to hold on any suitable possibility measure. A logical calculus permits us to specify a partial (qualitative) possibility measure and derive information implicit in the specification. We can reason without requiring complete information. One such possibilistic logic is developed in (Fariñas and Herzig 1991).

Others have provided logics in which logical constraints on probabilities can be specified in an analogous fashion (see, e.g., (Bacchus 1990; Frisch and Haddawy 1988), though these retain the quantitative aspects of probabilities).

Given the nature of necessity and possibility measures, the connection to modal logics is also of great interest (Dubois and Prade 1988) since the latter are typically put forth as representation systems for notions of possibility and necessity. We will present two modal logics, CO and CO*, in which we can faithfully represent the notions of qualitative necessity and possibility. These representations will respect the essential qualities of possibility and necessity measures. The expressive power we need to capture possibilistic logic is achieved with two modalities: the usual $\Box$, corresponding to truth at accessible worlds; and the less standard $\overleftarrow{\Box}$, expressing truth at *inaccessible* worlds. We note that, in contrast to many multimodal logics used in knowledge representation, the additional modality carries no excess semantical baggage. Our semantics is based on the usual Kripke structures for monomodal logics, the added modal operator increasing only our ability to constrain the form of such structures. The correspondence does not use the (perhaps expected) mapping of qualitative necessity and possibility into the operators $\Box$ and $\Diamond$. However, we provide other operators, defined using $\Box$ and $\overleftarrow{\Box}$ that do capture these absolute notions.

Aside from demonstrating that simple modal logics can be used to express qualitative possibility, the embedding into CO and CO* also illustrates important connections to a number of other formalisms for defeasible reasoning (which have also been mapped onto our logics). Some of these include conditional approaches to default reasoning (Lehmann 1989), $\varepsilon$-semantics (Pearl 1988), belief revision (Gärdenfors 1988), counterfactual logics (Lewis 1973) and autoepistemic logic (Moore 1985; Levesque 1990). In the next section we discuss qualitative possibility and present the logics CO and CO*. We show how these logics may be used to represent qualitative possibility. In Section 3, we examine the connections between possibilistic logic and some other systems for defeasible reasoning. Of particular interest is the fact we can define a conditional for default reasoning in terms of our possibility logics that is identical to Pearl's (1988) conditional for $\varepsilon$-semantics. This rela-tionship has been examined by Dubois and Prade (1991a), but our formulation has independent motivation, and lends itself to a complete calculus of conditionals. Furthermore, the expressive power of our logics allows us to express important properties that cannot be stated otherwise.

## 2  A Modal Representation of Possibility

### 2.1  Possibilistic Logic

Possibilistic logic has been developed to a great extent by Dubois and Prade (see their (1988) for a survey). A *possibility measure* $\Pi$ maps the sentences of $L_{CPL}$ into the real interval $[0, 1]$. The value $\Pi(\alpha)$ is intended to represent the degree of possibility of $\alpha$. We take this to represent the amount of surprise associated with adopting $\alpha$ as an epistemic possibility. If $\Pi(\alpha) = 1$ there is no surprise (i.e., $\alpha$ is consistent with the agent's beliefs), while $\Pi(\alpha) = 0$ indicates that surprise is maximal (i.e., an agent would *never* adopt $\alpha$). A possibility measure must satisfy the following three properties:

(a) $\Pi(\top) = 1$
(b) $\Pi(\bot) = 0$
(c) $\Pi(A \vee B) = \max(\Pi(A), \Pi(B))$

A *necessity measure* $N$ is a similar mapping, associating with $\alpha$ a degree of necessity. We take $N(\alpha)$ to represent the amount of surprise associated with giving up belief in $\alpha$ (or the degree of *entrenchment* of $\alpha$ in a belief set; see Section 4). If $N(\alpha) = 1$ then $\alpha$ is fully entrenched and can never be given up, while $N(\alpha) = 0$ indicates that $\alpha$ is not believed at all. Naturally, the degree of surprise associated with giving up a belief $\alpha$ should be related to the degree of surprise in accepting $\neg\alpha$ as an epistemic possibility, for giving up $\alpha$ is just accepting $\neg\alpha$ as possible. Indeed, one may define necessity measures using the identity

$$N(\alpha) = 1 - \Pi(\neg\alpha).$$

*Qualitative necessity measures* are discussed in (Dubois and Prade 1991b; Fariñas and Herzig 1991). Postulates are proposed constraining the qualitative relationship $\alpha \geq_N \beta$, which is read as "$\alpha$ is at least as necessary as $\beta$." If we define $\alpha \geq_N \beta$ to be true just when $N(\alpha) \geq N(\beta)$ for any necessity measure $N$, then $\geq_N$ will satisfy the postulates for qualitative necessity (in finite settings), and these relations are the only ones that can be so-defined (Dubois 1986). A qualitative necessity ordering is any ordering satisfying these postulates:[1]

(N1) $A \geq_N A$
(N2) $A \geq_N B$ or $B \geq_N A$
(N3) If $A \geq_N B$ and $B \geq_N C$ then $A \geq_N C$
(N4) $\top >_N \bot$
(N5) $\top \geq_N A$ for all $A$
(N6) If $B \geq_N C$ then $A \wedge B \geq_N A \wedge C$ for all $A$

*Qualitative possibility* is defined by related postulates, with $\alpha \geq_\pi \beta$ meaning $\alpha$ is at least as possible as $\beta$, or $\Pi(\alpha) \geq \Pi(\beta)$. The relationship between these qualitative relationships can be given as $A \geq_N B$ iff $\neg B \geq_\pi \neg A$.

Fariñas and Herzig (1991) have axiomatized this notion with a logic called *qualitative possibility logic* (QPL), in which the relation $\geq_\pi$ is incorporated as a conditional connective. They also make an initial attempt to develop a modal theory of possibility that uses only unary modal operators in place of the conditional $\geq_\pi$. Unfortunately, the resulting logic PL requires an infinite set of modal operators, each corresponding to a unique member of the measure set for $\Pi$. Semantically, each operator is evaluated with respect to a separate accessibility relation. This certainly permits the expression of qualitative properties like $\alpha \geq_\pi \beta$, but doesn't seem to reflect the qualitative nature of QPL or other qualitative postulates. In particular, there is no modal operator corresponding to (some degree of) possibility or necessity. This appears to be the first logical axiomatization of qualitative possibility and, as such, provides many of the advantages we expect of a logical calculus. Furthermore, they show QPL is equivalent to Lewis's (1973) logic VN.

### 2.2  The Logics CO and CO*

We wish to provide a possible worlds semantics for qualitative possibility theory, taking our models to consist of a set $W$ of possible worlds and a binary accessibility relation $R$ over $W$. Intuitively, $W$ is the set of situations an agent considers possible. We do not intend this to represent epistemic possibility, for there will be worlds among this set that are inconsistent with an agent's beliefs. Rather, these are the set of worlds an agent could *possibly* consider adopting, even if it changed its mind about certain beliefs it currently possesses. For example, $W$ could be the set of physically or logically possible worlds (for an agent).

We take $R$ to be a ranking of these worlds according to their *degree of possibility* or plausibility, the extent to which an agent is *willing* to accept these worlds as epistemically possible, or consistent with its beliefs. When $wRv$ we intend that $v$ is *at least as possible* as $w$. Intuitively, when $v$ is more possible than $w$ we can think of $v$ as being "more consistent" with an agent's current beliefs than $w$; or think of $v$ as a preferable, more plausible alternative state of affairs for an agent to adopt should it be forced to choose between the two.

We take as minimal requirements that $R$ be reflexive and transitive.[2] Another requirement we adopt in this paper is that of connectedness. In other words, any two states of affairs must be comparable according to their degree of possibility. If neither is more possible than the other, then they are equally possible.

If we intend the possibility ranking to respect an agent's current belief set, then it ought to be the case that the max-imally possible worlds in this ranking be precisely those the agent considers not merely possible, but *epistemically possible* (i.e., those worlds consistent with its beliefs). Although we do not need to enforce this constraint to deal with possibilistic logic, we will discuss how this can be expressed in Section 3, and how beliefs are related to possibility measures.

---

[1] For any ordering we propose (e.g. $\geq_N$), the corresponding relations $\leq$, $<$ and $>$ are defined in the standard way.

[2] In (Boutilier 1992a) we develop this minimal logic called CT4O in the context of belief revision.

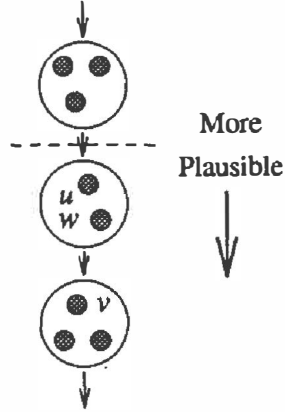

Figure 1: A CO-model

We now define a modal language with which we can express qualitative notions of possibility and necessity. Our language **L** will be formed from a denumerable set **P** of propositional variables, together with the connectives $\neg$, $\supset$, $\Box$ and $\overleftarrow{\Box}$. The connectives $\wedge$, $\vee$ and $\equiv$ are defined in terms of these in the usual way. We use $\top$ and $\bot$ to denote the identically true and false propositions, respectively. We denote by $\mathbf{L}_{CPL}$ the propositional sublanguage of **L**.

**Definition (Boutilier 1991)** A *CO-model* is a triple $M = \langle W, R, \varphi \rangle$, where $W$ is a set (of possible worlds), $R$ is a transitive, connected[3] binary relation on $W$ (the accessibility relation), and $\varphi$ maps **P** into $2^W$ ($\varphi(A)$ is the set of worlds where $A$ is true).

A CO-model consists of a set of *clusters* of possible worlds totally-ordered by $R$, where a cluster is a set of mutually accessible, or equally possible, worlds. In Figure 1 each large circle represents such a cluster of worlds and the (transitive closure of) arrows indicate accessibility between clusters. So $w$ can see both $u$ and $v$ ($wRu$ and $wRv$) but $v$ cannot see $w$. All worlds above the dashed line are inaccessible to $w$, while all worlds below are accessible. Let $M = \langle W, R, \varphi \rangle$ be a CO-model, with $w \in W$. The truth of a formula $\alpha$ at $w$ in $M$ is defined (for the interesting cases) as:

1. $M \models_w \Box \alpha$ iff for each $v$ such that $wRv$, $M \models_v \alpha$.

2. $M \models_w \overleftarrow{\Box} \alpha$ iff for each $v$ such that not $wRv$, $M \models_v \alpha$.

The sentence $\Box \alpha$ is true at $w$ if $\alpha$ holds at all worlds accessible to $w$. Given our reading of $R$, this means $\alpha$ must be true at all worlds that have a degree of possibility at least equal to that of $w$. The sentence $\overleftarrow{\Box} \alpha$, in contrast, holds when $\alpha$ is true at all inaccessible worlds, those strictly "less possible" than $w$. While the standard $\Box$ can force certain (classes of) worlds to be inaccessible, $\overleftarrow{\Box}$ can force certain worlds to be accessible. To illustrate the expressive power of CO, consider again Figure 1. If $w$ satisfies $\Box A$, this means no $\neg A$-worlds can be accessible to $w$. This forces all such worlds to be *inaccessible*. If the same world also satisfies $\overleftarrow{\Box} \neg A$, this means no $A$-worlds can be inaccessible, so all such worlds must be *accessible*. This type of constraint cannot be enforced using $\Box$ alone. When $w$ satisfies $\Box A \wedge \overleftarrow{\Box} \neg A$, essentially a line is drawn across the structure (as in Figure 1), all worlds above it satisfying $\neg A$ and all below it satisfying $A$.

We can define several new connectives as follows: $\Diamond \alpha \equiv_{df} \neg \Box \neg \alpha$; $\overleftarrow{\Diamond} \alpha \equiv_{df} \neg \overleftarrow{\Box} \neg \alpha$; $\overleftrightarrow{\Box} \alpha \equiv_{df} \Box \alpha \wedge \overleftarrow{\Box} \alpha$; and $\overleftrightarrow{\Diamond} \alpha \equiv_{df} \Diamond \alpha \vee \overleftarrow{\Diamond} \alpha$. It is easy to verify that these connectives have the following truth conditions: $\Diamond \alpha$ ($\overleftarrow{\Diamond} \alpha$) is true at a world if $\alpha$ holds at some accessible (inaccessible) world; $\overleftrightarrow{\Box} \alpha$ ($\overleftrightarrow{\Diamond} \alpha$) holds iff $\alpha$ holds at all (some) worlds, whether accessible or inaccessible. Validity is defined in a straightforward manner, a sentence $\alpha$ being *CO-valid* ($\models_{CO} \alpha$) just when every CO-model $M$ satisfies $\alpha$ at every world.

**Definition (Boutilier 1991)** The conditional logic CO is the smallest $S \subseteq \mathbf{L}$ such that $S$ contains CPL (and its substitution instances) and the following axiom schemata, and is closed under the following rules of inference:

**K** $\Box(A \supset B) \supset (\Box A \supset \Box B)$
**K'** $\overleftarrow{\Box}(A \supset B) \supset (\overleftarrow{\Box} A \supset \overleftarrow{\Box} B)$
**T** $\Box A \supset A$
**4** $\Box A \supset \Box \Box A$
**S** $A \supset \overleftarrow{\Box} \Diamond A$
**H** $\overleftrightarrow{\Diamond}(\Box A \wedge \overleftarrow{\Box} B) \supset \overleftarrow{\Box}(A \vee B)$
**Nec** From $A$ infer $\overleftrightarrow{\Box} A$.
**MP** From $A \supset B$ and $A$ infer $B$.

Provability and derivability are defined in the usual way, in terms of theoremhood (Hughes and Cresswell 1984).

**Theorem 1 (Boutilier 1991)** $\vdash_{CO} \alpha$ iff $\models_{CO} \alpha$.

We often want to ensure that all logically possible worlds are taken into consideration in our models (for instance in the context of belief revision (Boutilier 1992c) or autoepistemic reasoning (Levesque 1990; Boutilier 1992b)). In our current setting, we can think of this as ensuring that every logically possible world is assigned some positive degree of possibility. For this purpose we introduce the logic CO*, which is based on the class of CO-models in which all propositional valuations are represented (see also (Levesque 1990)).

**Definition (Boutilier 1991)** CO* is the smallest extension of CO closed under all rules of CO and containing the following axioms:

---
[3]$R$ is (totally) connected if $wRv$ or $vRw$ for any $v, w \in W$ (this implies reflexivity).

**LP** $\tilde{\Diamond}\alpha$        for all satisfiable propositional $\alpha$.

**Definition (Boutilier 1991)** A *CO\*-model* is any CO-model $M = \langle W, R, \varphi \rangle$, such that
$\{f : f \text{ maps } \mathbf{P} \text{ into } \{0, 1\}\} \subseteq \{w^* : w \in W\}$.[4]

**Theorem 2 (Boutilier 1991)** $\vdash_{CO*} \alpha$ *iff* $\models_{CO*} \alpha$.

### 2.3 A Modal Account

We now take up the task of providing a simple modal account of qualitative possibility and necessity. Recall that a CO-structure orders worlds according to their degree of possibility. This is much like another formulation of possibilistic logic in terms of *possibility distributions*. A distribution $\pi$ assigns to each world a degree of possibility from the interval $[0, 1]$. This too can be viewed as a ranking of worlds, with $w$ being at least as possible as $v$ just when $\pi(w) \geq \pi(v)$, corresponding precisely to $vRw$ in a CO-model. A distribution determines a possibility measure $\Pi$ via the following relationship:

$$\Pi(A) = \max\{\pi(w) : w \models A\}.$$

In other words, the degree of possibility of $A$ is just that of the *most* possible $A$-world.

In our qualitative setting we need not determine the absolute possibility of $A$, merely the relative possibility of pairs of sentences. Quantitatively, $A$ is at least as possible as $B$ iff the world of maximal possibility satisfying $A$ (say $w$) is no less possible than the most possible $B$-world (say $v$); that is, $\pi(w) \geq \pi(v)$. In a CO-structure this means $vRw$. So if $A$ is at least as possible as $B$, then it must be the case that the *minimal* $B$-world in relation $R$ can see the minimal $A$-world.[5] But, since $v$ is a minimal $B$-world, all $B$-worlds see $v$; therefore, any $B$-world can see $w$ (since $R$ is transitive and connected). More generally, whenever $A$ is at least as possible as $B$, *any* $B$-world can see *some* $A$-world. In our bimodal language this is expressed as $\tilde{\Box}(B \supset \Diamond A)$: whenever $B$ holds, there is some more plausible world satisfying $A$. We refer to this as a *plausibility ordering* (for reasons discussed in the next section).

**Definition** Let $M$ be a CO-model. The *plausibility ordering* determined by $M$ is $\leq_{PM}$, given by

$A \leq_{PM} B$ iff $M \models \tilde{\Box}(B \supset \Diamond A)$.

$A$ is *at least as plausible* as $B$ iff $A \leq_{PM} B$.[6]

---

[4] For all $w \in W$, $w^*$ is defined as the map from $\mathbf{P}$ into $\{0, 1\}$ such that $w^*(A) = 1$ iff $w \in \varphi(A)$; in other words, $w^*$ is the valuation associated with $w$.

[5] We are speaking loosely here, of course. World $w$ is a *minimal* $A$-world iff $vRw$ for all $A$-worlds $v$. (Worlds *minimal* in $R$ have *maximal* plausibility.) In this case $w$ is "less than" any such $v$ according to $R$, but (on our interpretation of $R$) has "greater" possibility than $v$. Nothing about CO-structures presupposes the existence of minimal $A$-worlds for any $A$, nor need they be unique when they do exist. We use this manner of speaking for illustrative purposes; but nothing of a technical nature depends on this.

[6] We use $\leq_{PM}$ to indicate greater plausibility rather than $\geq_{PM}$ to remain consistent with (Grove 1988; Boutilier 1992a) and other papers (where this operator is related to other concepts).

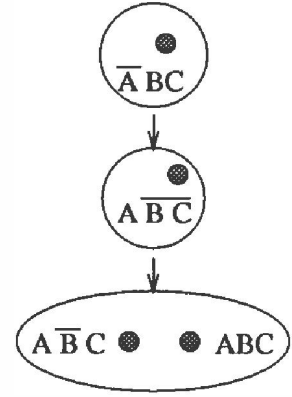

Figure 2: Entrenchment and Plausibility

The dual of such a relationship is qualitative necessity, and we refer to this ordering as an *entrenchment ordering* (again, explained in the next section).

**Definition** Let $M$ be a CO-model. The *entrenchment ordering* determined by $M$ is $\leq_{EM}$, given by

$B \leq_{EM} A$ iff $\neg B \leq_{PM} \neg A$.

$A$ is *at least as entrenched as* $B$ iff $B \leq_{EM} A$.

It is easy to see that $B \leq_{EM} A$ iff $M \models \tilde{\Box}(\neg A \supset \Diamond \neg B)$.

Figure 2 shows a CO-model where $A$, $B$, $\neg B$ and $C$ are each more plausible than $\neg A$. Every world where $\neg A$ holds is strictly less plausible than some world where these other propositions hold. We also see that $A \wedge \neg C$ is more plausible than $\neg A \wedge C$. $A$ and $C$ are equally (and maximally) plausible, yet $A$ is more entrenched than $C$. This is due to the fact that as we "move up" from the bottom cluster, we find a $\neg C$-world before a $\neg A$-world. $\neg C$ is more readily "accepted" than $\neg A$, so $C$ is less entrenched. Notice, since there are no worlds satisfying (say) $\neg A \wedge \neg B$ in the model, we judge all such worlds to have no plausibility, $\pi(w) = 0$. Correspondingly, according to our definition of $\leq_{PM}$, $\neg A \wedge \neg B$ is (strictly) less plausible than any sentence satisfied in the model. Furthermore, every sentence in the language is at least as plausible as $\neg A \wedge \neg B$. That is, $\alpha \leq_{PM} \neg A \wedge \neg B$ for all $\alpha$ (including $\alpha = \bot$). Thus, we see that $\Pi(\neg A \wedge \neg B) = 0$. As mentioned above, one should think of the $R$-minimal worlds in a model (those with maximal plausibility) as representing the epistemic state of the agent in question. In other words, each minimal world is epistemically possible. In this example, we consider the two lowest worlds to be those consistent with the agent's beliefs, while all other worlds violate some belief. Here, the agent believes $A \wedge C$. In the next section we will see how belief can be expressed at the object level.

We can show two key results concerning this model of qualitative possibility.

**Theorem 3** *Any entrenchment ordering determined by a CO-model $M$ is a qualitative necessity ordering satisfying postulates (N1) through (N6).*

**Theorem 4** *For any qualitative necessity ordering $\geq_N$ there is a CO-model M determining the corresponding entrenchment ordering: $A \geq_{EM} B$ iff $A \geq_N B$.*

These results show that necessity orderings and entrenchment orderings determined by CO are exactly the same. It immediately follows that the space of plausibility orderings determined by CO-models corresponds precisely to the set of qualitative possibility orderings. The first theorem is easy to verify using the definition of $A \leq_{EM} B$ and the logical properties of CO. The second theorem can be shown as follows: we consider the corresponding qualitative possibility measure $\geq_\pi$ and construct a CO-model that determines the plausibility ordering (it is easy to see that the dual entrenchment and necessity orderings will then be equivalent as well). We can construct a CO-model $M$ using as worlds the maximal consistent sets of $\mathbf{L}_{CPL}$ (which determine the obvious valuations). We define *cuts* on the language as any set $\mathcal{C} \subseteq \mathbf{L}_{CPL}$ such that if $A \in \mathcal{C}$ and $A \geq_\pi B$ then $B \in \mathcal{C}$ (see Grove (1988), who uses this technique). We exclude from $M$ all worlds that intersect only the minimal cut, $\{A : \bot \geq_\pi A\}$, and define accessibility as follows:

> $wRv$ iff every cut $\mathcal{C}$ that intersects $v$ also intersects $w$ (i.e., $v \cap \mathcal{C} \neq \emptyset$ implies $w \cap \mathcal{C} \neq \emptyset$).

It is easy to verify that $M$ is a CO-model. We can also show that $M \models \tilde{\Box}(B \supset \Diamond A)$ iff $A \geq_\pi B$. We sketch this briefly, assuming $B >_\pi \bot$. Let $w$ be a world containing $B$ and $\mathcal{C}$ be the "smallest" cut that intersects $w$. If $A \geq_\pi B$ the properties of $\geq_\pi$ ensure that: a) if $A$ is consistent with the negation of all sentences in $\mathcal{C}$ then there is a maximal consistent set $v$ containing $A$ disjoint from $\mathcal{C}$; b) if $A$ is inconsistent with this set it is easy to verify that $A$ is in $\mathcal{C}$ but that $A$ is in no smaller cut than $\mathcal{C}$. In either case $wRv$ for some $A$-world $v$ and $M \models \tilde{\Box}(B \supset \Diamond A)$. The converse is quite clear.

As discussed by Dubois (1986), for finitary languages a qualitative possibility measure is compatible with a mapping of sentences into the interval $[0, 1]$ iff the mapping is a possibility measure (compatibility means $A \geq_\pi B$ iff $\Pi(A) \geq \Pi(B)$). Hence, plausibility orderings in CO are precise qualitative counterparts of possibility measures, and entrenchment orderings correspond to necessity measures. Our treatment of plausibility and entrenchment generalizes that of Dubois and Prade by permitting orderings on infinite languages (with CO-models serving as adequate representation structures).

## 3 Beliefs and Conditionals

### 3.1 Beliefs

We have seen how to express qualitative possibility and necessity measures using two modal operators. More importantly, the relationship to CO allows us to exhibit connections between possibility theory and other forms of defeasible reasoning. We begin by explaining the choice of terminology of the last section. The ordering $\leq_{EM}$ determined by some CO-model turns out to be an *expectation ordering* in the sense of Gärdenfors and Makinson (1991). These themselves are weakenings of orderings of *epistemic entrenchment* (Gärdenfors 1988), which are intended to represent the degree of certainty of elements of a belief set $K$. For any (deductively closed) belief set $K$ an entrenchment ordering $\leq_E$ ($A \leq_E B$ means $B$ is at least as entrenched as $A$) satisfies the postulates

**(E1)** If $A \leq_E B$ and $B \leq_E C$ then $A \leq_E C$

**(E2)** If $A \vdash B$ then $A \leq_E B$

**(E3)** If $A, B \in K$ then $A \leq_E A \wedge B$ or $B \leq_E A \wedge B$

**(E4)** If $K \neq Cn(\bot)$ then $A \notin K$ iff $A \leq_E B$ for all $B$

**(E5)** If $B \leq_E A$ for all $B$ then $\vdash A$

Dubois and Prade (1991b) show a partial correspondence between qualitative necessity and entrenchment. For any necessity ordering $\geq_N$, they define the set of beliefs associated with $\geq_N$ to be

$$K = \{\alpha : \alpha >_N \bot\}$$

Assuming that $N(\bot) = 0$ for any necessity measure used to "generate" the necessity ordering, this means $N(\alpha) > 0$. Thus, $\alpha$ is believed just when it has *some* degree of necessity. Entrenchment and qualitative necessity correspond if we ignore (N4) and (E5). Entrenchment fails to satisfy (N4) only when every sentence is equally entrenched (including $\bot$); that is, when we are dealing with the inconsistent belief set. We will ignore this case and assume that entrenchment orderings are nontrivial, satisfying (N4).[7]

Qualitative necessity fails to satisfy (E5) because certain nontautologous beliefs are allowed to be certain or completely necessary (i.e., $N(\alpha) = 1$). In general, entrenchment orderings determined by CO-models will not satisfy (E5). But if we consider only CO*-models, every logically consistent $\alpha$ has some degree of possibility, and every contingent sentence will be less certain than $\top$. Thus (E5) is satisfied by the *full* qualitative necessity ordering determined by any CO*-model.

**Theorem 5** (Boutilier 1992b) *Any entrenchment ordering determined by a CO*-model satisfies (E1) – (E5).*

**Theorem 6** (Boutilier 1992b) *For any entrenchment ordering $\leq_E$ there is a CO*-model M determining the corresponding entrenchment ordering: $A \leq_{EM} B$ iff $A \leq_E B$.*

Of course, the real reason for examining logics of qualitative necessity and possibility is to provide a method of expressing and reasoning with qualitative constraints on necessity and possibility (i.e., premises) without relying on complete knowledge of, say, a possibility ordering or measure. Given certain constraints we can determine through logical deduction what must be true in all measures or orderings satisfying these constraints. The expressive power of CO and CO* can also be used to capture notions that are not amenable to direct analysis using a simple language of qualitative necessity or possibility (e.g., Fariñas and Herzig's QPL).

---

[7]We can capture the trivial ordering by considering the empty "CO-model" as a model for entrenchment. Axiomatically we can express the ordering using the inconsistent theory $\{\bot\}$.

Naturally, we'd like to express relationships of qualitative possibility. In QPL one may assert $A \geq_\pi B$, while in CO we say $\vec{\Box}(B \supset \Diamond A)$ to indicate that $A$ is as possible as $B$. Absolute concepts such as belief, disbelief, possibility and necessity are important as well. These can also be asserted in QPL; for example, "$A$ is believed" is just $\top >_\pi \neg A$. In CO, these modalities can be expressed as follows:

- When $\alpha$ is *believed* it must have some degree of necessity ($N(\alpha) > 0$). In CO this is expressible as $\overset{\leftrightarrow}{\Diamond}\Box\alpha$. This is true exactly when there is some point is the model where $\alpha$ holds at all more plausible worlds. In particular, such a point exists when $\alpha$ holds at the *most* possible worlds (assuming such a limit). We define a modality for belief in this way, $\mathsf{B}\alpha$ denoting $\overset{\leftrightarrow}{\Diamond}\Box\alpha$. The model in Figure 2 satisfies $\mathsf{B}A$ and $\mathsf{B}C$. It is easy to verify that CO satisfies the axioms of the belief logic weak S5 when $\mathsf{B}$ is taken as a modal operator. Thus, qualitative possibility respects some reasonable conditions on the beliefs it determines. Some of the more notable properties are the introspective axioms, which are valid in CO:

$$\mathsf{B}\alpha \supset \mathsf{BB}\alpha \quad \text{and} \quad \neg\mathsf{B}\alpha \supset \mathsf{B}\neg\mathsf{B}\alpha$$

Notice that $\mathsf{B}\alpha$ and $\mathsf{B}\neg\alpha$ are mutually inconsistent, but $\alpha$ and $\mathsf{B}\neg\alpha$ are not.

- *Disbelief* is expressed as $\neg\mathsf{B}\alpha$. This is true just when $N(\alpha) = 0$, or $\Pi(\neg\alpha) = 1$. Notice that $\neg\mathsf{B}\alpha$ and $\neg\mathsf{B}\neg\alpha$ are mutually consistent. The model in Figure 2 satisfies $\neg\mathsf{B}B$ and $\neg\mathsf{B}\neg B$.

- If $\alpha$ has *some degree of possibility* ($\Pi(\neg\alpha) > 0$), $\neg\alpha$ cannot be certain. This holds exactly when $\overset{\leftrightarrow}{\Diamond}\alpha$ is verified ($\alpha$ is true at some world with a nonzero degree of possibility).

- Finally, $\alpha$ is *completely necessary* ($N(\alpha) = 1$) exactly when $\overset{\leftrightarrow}{\Box}\alpha$ holds. The model in Figure 2 satisfies $\overset{\leftrightarrow}{\Box}(A \vee B)$ since $A \vee B$ holds at each world (it is completely necessary). $\neg A \wedge \neg B$ is accorded no possibility at all. *Some degree of necessity* is assigned to $\alpha$ ($N(\alpha) > 0$) just when it is believed; that is, $\mathsf{B}\alpha$ is true. So, in the example, $B$ and $\neg B$ have a necessity measure of zero (since neither is believed). $A$ and $C$ are accorded some (less than absolute) degree of necessity, with $A$ being more necessary (or entrenched) than $C$.

CO and CO* are much stronger than this. In particular, these logics allow us to express the concept of *only knowing*. To only know (or only believe) a sentence $\alpha$ is to believe $\alpha$ and to believe nothing more than is required by $\alpha$ (Levesque 1990). For example, given a (finite) knowledge base $KB$, we usually intend that $KB$ is all that is believed.[8] If $KB \models \alpha$ then $\alpha$ is believed; if $KB \not\models \alpha$ then $\alpha$ is not believed. In the usual epistemic logics, merely asserting $\mathsf{B}(KB)$ does not carry this force. Indeed, $\mathsf{B}(KB)$ does not preclude the possibility of $\mathsf{B}\alpha$ when $KB \cup \{\alpha\}$ is consistent, even if $KB \not\models \alpha$. To express that the sentences in $KB$ are believed in QPL, we need only assert that $\top >_\pi KB$ (or that $N(KB) > 0$ in a quantitative setting). But there are no convenient and systematic means of asserting that these are the *only* beliefs, or that these are the only sentences that have some positive degree of necessity.

---

[8]We will often use $KB$ as if it were the conjunction of its elements (a sentence). For a fuller discussion of only knowing see (Levesque 1990; Boutilier 1992b).

In CO, we can express the fact that $KB$ is all that is believed using the sentence

$$O(KB) \equiv_{df} \overset{\leftrightarrow}{\Box}(KB \supset (\Box KB \wedge \overset{\leftrightarrow}{\Box}\neg KB))$$

Typically, we consider only CO*-models when discussing "all that is known," for this terminology suggests that no logical possibilities should be excluded from consideration. When $KB$ is believed, only $KB$-worlds can be accepted as epistemically possible. When $KB$ is all that is believed, not only should $KB$ be believed, but *every* $KB$-world should be accepted as epistemically possible. If some world is not accepted, then there should be some belief that excludes this world from consideration, some belief falsified by that world. If a world satisfies $KB$, there is no such belief when $KB$ is all that is believed.

For purely propositional $KB$ we have that

$$O(KB) \models_{CO_*} \mathsf{B}\alpha \text{ iff } KB \models \alpha$$
$$O(KB) \models_{CO_*} \neg\mathsf{B}\alpha \text{ iff } KB \not\models \alpha$$

(see (Boutilier 1992b) for details). In particular, the only sentences assigned a degree of necessity greater than 0 are those entailed by $KB$. In a natural and convenient fashion we can summarize what would require an infinite set of sentences in QPL, (or an unwieldy number for finite languages). The model in Figure 2 satisfies $O(A \wedge C)$, assuming a language with only three atoms (and ignoring the fact that this is not a CO*-model). Thus, $O(A \wedge C)$ is a concise way of expressing that *only* the consequences of $A \wedge C$ are assigned a positive degree of necessity. Again, this is crucial since, when one specifies some knowledge base $KB$, it is usually intended that only those sentences derivable from $KB$ are believed.

The expressive power of CO goes beyond this, however. Nothing prevents the occurrence of nonpropositional sentences in $KB$. We can have belief sentences in $KB$, and even sentences of an autoepistemic nature. In fact, in (Boutilier 1992b) we show that CO* subsumes autoepistemic logic. So we can think of CO* as adding to qualitative possibility logic the ability to express autoepistemic reasoning. With this connection, of course, degrees of possibility or entrenchment can be interpreted as generalizing autoepistemic logic as well.

### 3.2 Conditionals

CO has been used as a conditional logic for representing default rules. In (Boutilier 1990; Boutilier 1991) we define a conditional connective $\Rightarrow$, reading $A \Rightarrow B$ as "$A$ normally implies $B$." We can show that CO*, used in this way, captures Lehmann's (1989) rational consequence relations and Pearl's (1988) $\varepsilon$-semantics. This connective can be related to our plausibility orderings as follows:

**Theorem 7** *Let $M$ be a CO-model. Then $M \models A \Rightarrow B$ iff $A \wedge B <_{PM} A \wedge \neg B$ or $M \models \overset{\leftrightarrow}{\Box} \neg A$.*

In other words, $A \Rightarrow B$ holds just when $A \wedge B$ is more possible than $A \wedge \neg B$ (or $A$ is impossible, $\Pi(A) = 0$). With a little simplification of these conditions, we obtain the definition of the connective $\Rightarrow$ as presented by Boutilier (1990; 1991)

$$A \Rightarrow B \equiv_{df} \overset{\leftrightarrow}{\Box} \neg A \vee \overset{\leftrightarrow}{\Diamond}(A \wedge \Box(A \supset B))$$

Thus we can define an inferential relation on conditional sentences (or default rules) using qualitative possibility, and it will be equivalent to a number of other systems of defeasible inference. These include the following systems (whose relationship to CO is explored in the corresponding references): Pearl's (1988) $\varepsilon$-semantics (Boutilier 1992d; Boutilier 1990); Lehmann's (1989) preferential and rational consequence relations (Boutilier 1990); the purely conditional fragment of Lewis's (1973) counterfactual logic VC (Boutilier 1992c). As discussed above, Gärdenfors and Makinson's (1991) notion of *expectation inference*, based on their expectation orderings, also corresponds to this sort of conditional possibilistic inference (ignoring the trivial expectation ordering), a connection they point out. We note that most of these equivalences rely on our specification of qualitative necessity and possibility in terms of nonfinite languages.

**Example** Let $A$, $S$, $E$ stand for "adult", "grad student" and "employed", respectively, and consider the following set of premises (a standard example from the default reasoning literature):

$$\{A \Rightarrow E, S \Rightarrow A, S \Rightarrow \neg E\}$$

Our conditionals are exception-allowing since $A \wedge \neg E$ is consistent with this theory. Preference for more specific defaults is automatically incorporated into the definition of $\Rightarrow$ as well. From this theory we can derive $S \wedge A \Rightarrow \neg E$ using consequence in CO, but we cannot derive $S \wedge A \Rightarrow E$. Also derivable are constraints on permissible possibility assignments, for example, it must be that $A >_\pi S$. It is more plausible that someone is simply an adult than a grad student.

The relationship with $\varepsilon$-semantics holds particular interest since its semantic foundations rely on probabilistic notions. By presenting qualitative possibility and $\varepsilon$-semantics within our modal framework we can show the following equivalence. We assume $\alpha \rightarrow \beta$ is some abstract conditional, being interpreted either as $\Rightarrow$ in CO or as a default *rule* in the sense of $\varepsilon$-semantics (we assume $\alpha$ is satisfiable).

**Theorem 8** *Let $T$ be a finite conditional theory consisting of sentences of the form $\alpha \rightarrow \beta$, where $\alpha, \beta \in \mathbf{L}_{CPL}$. Then $T \vdash_{CO} \alpha \rightarrow \beta$ iff $T$ $\varepsilon$-entails $\alpha \rightarrow \beta$.*

This follows from the results of (Boutilier 1990) and (Kraus, Lehmann and Magidor 1990), but a direct proof of this result is given in (Boutilier 1992a).

This connection has also been examined by Dubois and Prade (1991a). Using CO as the intermediate framework between $\varepsilon$-semantics and qualitative possibility allows us to see the underlying semantic commonality in these systems. Adams's (1975) construction for determining the consistency of theories of statements about arbitrarily high probabilities can be interpreted as ranking possible worlds according to their degree of probability. Given this ranking, it is easy to ensure that the conditional probabilities of the statements in the theory are as high as they need to be. But this ranking can also be construed as a simple CO-model. Our interpretation of CO-models in this paper equates this ranking with the degree of possibility of worlds. On either interpretation of the models, the same conclusions are derivable from simple conditional theories.

The results of Boutilier (1990) also show that $\varepsilon$-semantics can be modeled in the monomodal logic S4.[9] Thus for the purely conditional fragment of qualitative possibility theories, representation and inference can be performed using S4 (and conversely, if S4 is restricted to its simple "conditional" fragment).[10]

Once we allow boolean combinations of conditionals, it is not clear that the intuitions underlying Adams's approach remain viable. Our semantics for qualitative possibility is more compelling in this case. We must also contrast our approach with the model of conditional possibility adopted by Dubois and Prade (1991a). They provide a semantics for (some) boolean combinations of conditionals defined in terms of possibility measures. Unfortunately, they equate the "weak negation" of a conditional $\neg(A \Rightarrow B)$ with the "strong negation" $A \Rightarrow \neg B$. This is certainly not the case on our definition of conditionals. Merely denying a conditional is not reason to accept that the antecedent justifies acceptance of the negation of the consequent. It should be quite reasonable to say "My door is not (normally) open or closed." In CO, the following is consistent:

$$\neg(A \Rightarrow B) \wedge \neg(A \Rightarrow \neg B).$$

Our extension of the conditional language is much more compelling in this respect.

## 4 Concluding Remarks

We have presented two modal logics for reasoning about orderings of qualitative necessity and possibility. The expressive power of CO and CO* can be used to express constraints on possibility measures in a natural and concise fashion (for example, through only knowing). As pointed out by a number of people, the numbers attached to propositions by possibility measures are perhaps of less importance than the ranking of the propositions. We are able to exploit this fact in developing a simple semantic account of qualitative possibility. This simple view allows us to exhibit the connection between possibility theory and a number of other forms of defeasible reasoning. Furthermore, these modal possibilistic logics provide a means of representing very general constraints on possibility measures, since we allow arbitrary boolean combinations of formulae.

---

[9]See also (Boutilier 1992a) for a more detailed presentation.

[10]We note that S4 structures are precisely CO-models without the requirement of connectedness. While this relaxation is not appropriate in general, simple conditional theories cannot express the distinction between the two types of structures. Thus the simple fragment of the (mono-) modal logic S4.3 (characterized by the class of connected, or CO, models) is also equivalent to these logics. See (Boutilier 1990) for details.

A number of avenues remain to be explored. By generalizing the logic CO, we can explore weaker types of possibilistic semantics. For example, by dropping the requirement of connectedness (obtaining the logic CT4O of (Boutilier 1992a)) we are in essence modeling partially ordered possibilistic measure sets. Though we have demonstrated or pointed to a number of connections to existing systems of defeasible reasoning, some of these remain to be explored in detail. A number of other interesting relationships are brought to light by this work as well. Possibilistic logic has strong ties to Shafer's belief functions (Dubois and Prade 1991b). This suggests a link to the forms of defeasible reasoning discussed in the last section, a connection we have yet to explore. We have also begun preliminary investigations into the relationship between our conditional approaches to default reasoning based on CO* and probabilistic systems of reasoning such as that of Bacchus (1990), which could possibly bring to light further connections to possibilistic logic.

## Acknowledgements

I would like to thank Luis Fariñas del Cerro, David Makinson, and David Poole for their valuable comments and suggestions, as well as the referees.